\documentclass{article}
\usepackage{spconf,amsmath,graphicx}
\usepackage{booktabs, multicol, multirow}
\usepackage{subfigure}
\usepackage{array}
\usepackage{color}

\pdfoutput=1 

\newcommand{\comment}[1]{\textcolor{black}{#1}}

\title{MULTIBIOMETRIC SECURE SYSTEM BASED ON DEEP LEARNING}
\name{Veeru Talreja, Matthew C. Valenti, and Nasser M. Nasrabadi}
\address{West Virginia University, Morgantown, USA}
\begin{document}
%
\maketitle
\begin{abstract}
In this paper, we propose a secure multibiometric system that uses deep neural networks and error-correction coding. We present a feature-level fusion framework to generate a secure multibiometric template from each user's multiple biometrics. Two fusion architectures, fully connected architecture and bilinear architecture, are implemented to develop a robust multibiometric shared representation. The shared representation is used to generate a cancelable biometric template that involves the selection of a different set of reliable and discriminative features for each user. This cancelable template is a binary vector and is passed through an appropriate error-correcting decoder to find a closest codeword and this codeword is hashed to generate the final secure template. The efficacy of the proposed approach is shown using a multimodal database where we achieve state-of-the-art matching performance, along with cancelability and security.
\end{abstract}
\begin{keywords}
Multibiometrics, template protection, Reed-Solomon code, secure-sketch, cancelable biometrics. 
\end{keywords}
\section{Introduction}
\label{sec:intro}

Multibiometric systems offer several advantages when compared to unibiometric systems, including better recognition accuracy, security, flexibility, population coverage and user convenience \cite{nandakumar_multibiometric_2008}. However, multibiometric systems have an increased demand for integrity and privacy because the system stores multiple biometric traits of each user. 

There has recently been a significant amount of research in secure biometrics, which can be grouped into two categories: \emph{biometric cryptosystem} and \emph{transformation based methods} \cite{rane_secure_biom_2013}. \emph{Fuzzy commitment} and \emph{secure sketch} are biometric cryptosystem methods and are usually implemented using error correcting codes and provide information-theoretic guarantees of security and privacy (e.g., \cite{Sutcu_2008_Feature_SW,Sutcu_2008_Feature_ECC,Sutcu_2007_Protecting,Juels_1999_Fuzzy,Juels_2002_Vault,Nandakumar_2007_Fingerprint_FV,Nagar_2008_Securing}). Cancelable biometrics use revocable and non-invertible user-specific transformations for distorting the enrollment biometric (e.g., \cite{ratha_canc_biom_2007,kong_analysis_2006,zuo_cancelable_2008,teoh_cancellable_2008}),  with the matching typically performed in the transformed domain. 

One important issue of a multibiometric secure system is that biometric traits being combined may not follow the same feature-level representation,  and it is difficult to characterize multiple biometric traits using compatible feature-level representations, as required by a template protection scheme \cite{nagar_multibiometriccryptosystems_2012}. To counter this issue there have been many fusion techniques for combining multiple biometrics in the past \cite { sutcu_secure_2007,nandakumar_multibiometric_2008,nagar_multibiometriccryptosystems_2012}. One of the possible techniques is to apply a separate template protection scheme for each trait followed by decision-level fusion. But such an approach may not be highly secure, since it is limited by the security of the individual traits. 

Motivated by the drawbacks of systems that protect biometric traits separately, we propose in this paper a \emph{multibiometric secure scheme}  for face and iris biometrics. This scheme is based on cancelable biometrics and forward error control (FEC) codes where the feature extraction for individual modalities and joint feature-level fusion are performed using a Deep Neural Network (DNN). The proposed approach involves the following steps:

1) The face and iris biometrics are converted into a common feature space by extracting domain-specific features using dedicated Convolutional Neural Networks (CNNs). These features are then fused with the help of a joint representation layer (fully connected layer or bilinear layer). 


2) The final fused feature vector is reduced in dimension to generate a multibiometric cancelable template by applying a feature selection process. 

3) The cancelable template is a binary vector that is within a certain distance from a codeword of an error-correcting code, which for purposes of illustration we assume is a Reed Solomon code.  By passing the template through an appropriate decoder, the closest codeword is found, and that codeword is hashed to generate the final secure template.   
\section{Multibiometric Secure Scheme}
\label{sec:arch}

\begin{figure}[t]
\centering
\includegraphics[width=8.50cm]{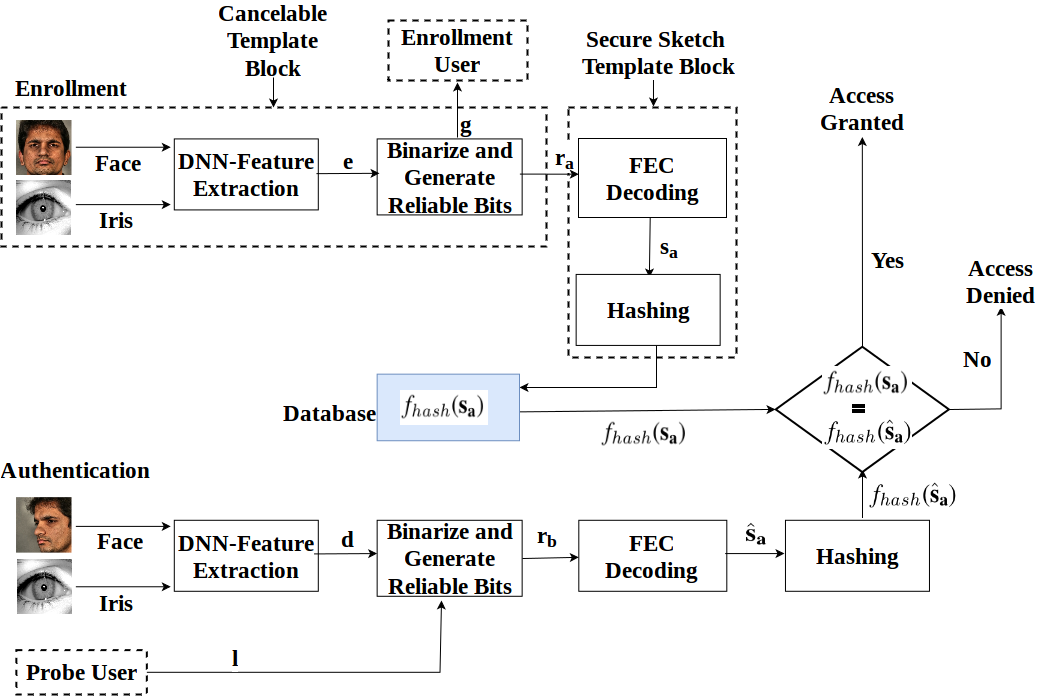}
\caption{Enrollment and Authentication Block Diagram.}\label{fig:enrol} 
\vspace{-0.60cm}
\end{figure}

The block diagram for the proposed framework is shown in Fig. \ref{fig:enrol}. There are two important blocks for this framework: \emph{Cancelable Template Block (CTB)} and \emph{ Secure Sketch Template  Block (SSTB)} which will be explained in this section. 
\vspace{-0.40cm}
 
\vspace{-0.30cm}





\subsection{Cancelable Template Block (CTB)}\label{subsec:CTB}
\begin{figure}[b]
\vspace{-0.30cm}
\centering
\includegraphics[width=8.5cm]{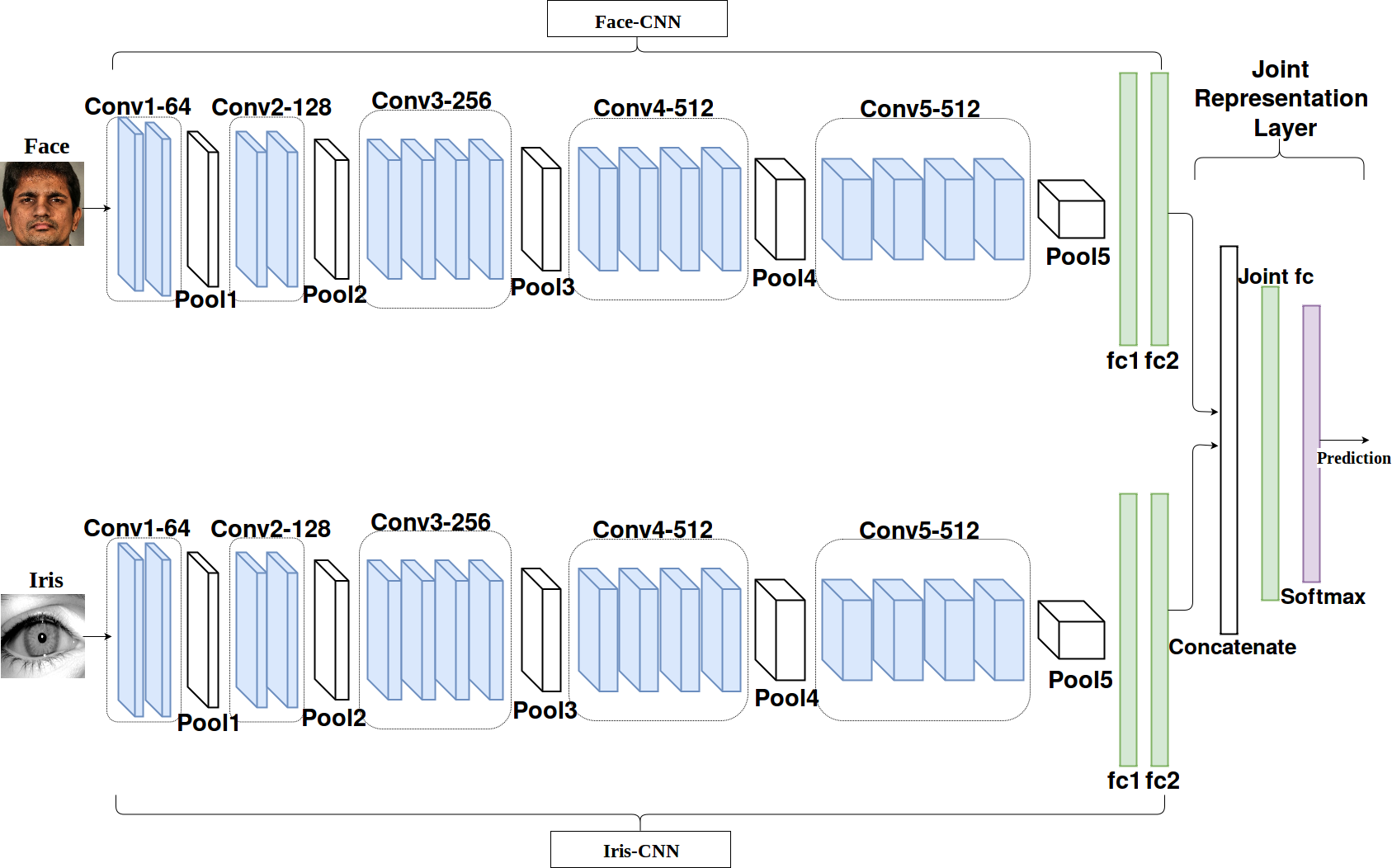}
\caption{Proposed Deep Neural Network (DNN) Fully Connected Architecture.}\label{fig:arch} 
\vspace{-0.60cm}
\end{figure}

The CTB contains the DNN based feature extraction and fusion components followed by the binarizing and reliable bit extraction stages, which generate a cancelable multibiometric template. The architecture for the DNN is shown in Fig. \ref{fig:arch}. The domain-specific layer of the DNN consists of a CNN for encoding the face ("Face-CNN") and a CNN for encoding the iris ("Iris-CNN"). For each CNN, we use VGG-19 \cite{simonyan_very_deep_2014} pre-trained on ImageNet \cite{deng_imagenet_2009} as our starting points followed by fine-tuning on different datasets (section \ref{subsec:DS}). 

The output feature vectors of the Face-CNN and Iris-CNN are fused in the joint representation layer. We have implemented two different architectures for the joint representation layer: Fully Connected architecture (FCA) and Bilinear architecture (BLA). In FCA (Fig.\ref{fig:arch}), the outputs of the Face-CNN and Iris-CNN are concatenated vertically and passed through a fully connected layer to fuse the iris and face features. In BLA, the outputs of the Face-CNN and Iris-CNN are combined using the matrix outer product of the face and iris feature vectors. For training the joint representation layer we adopt a two-step training procedure where we first train only the joint representation layer greedily with softmax for classification followed by fine tuning the entire model end-to-end using back-propagation at a relatively small learning rate. 

The output of the joint representation layer is a real valued shared multimodal feature vector for face and iris biometrics. This shared representation vector is binarized and user-specific reliable bits are extracted using a method that is similar to the method used by Kevenaar et.al. in \cite{kevenaar_face_2005}. This user-specific reliable bit vector forms the cancelable template. 
\subsection{Secure Sketch Template Block (SSTB)}\label{subsec:SSTB}

The SSTB contains the forward error correction (FEC) decoding stage followed by the hashing stage. The FEC decoding implemented in our framework is the equivalent of the secure-sketch template protection scheme. In the secure-sketch scheme, error control coding is applied on the biometric feature vector to generate a sketch which is stored as the template. In our work, the cancelable template generated from the CTB is considered to be the noisy codeword of $N$ symbols. This noisy codeword is decoded with a FEC decoder and the output of the decoder is the multibiometric sketch $\textbf{s}_\textbf{a}$ that corresponds to the codeword closest to the cancelable template. This multibiometric sketch $\textbf{s}_\textbf{a}$ is hashed and stored in the database. The authentication is also performed using FEC decoding with the same parameters to give an estimate of the multibiometric sketch $\hat{\textbf{s}}_\textbf{a}$ which is hashed and access is granted only if the hash matches the enrolled hash. We have implemented Reed-Solomon (RS) decoder for FEC decoding because we can exploit the maximum distance seperable (MDS) property of RS codes to manage the intra-class variability of the joint template.

This error control coding method can be explained in the context of standard-array decoding. A standard array is a two-dimensional array where the first row contains the codewords and every other row is coset of the code, which is found by adding an unused vector (error pattern) to all the codewords of the code. Decoding using the standard array is performed by decoding the received vector \textbf{r} to the codeword at the top of the column that contains \textbf{r}. The decoding region for a codeword is the column headed by that codeword. In our framework, the output of the CTB in both enrollment and authentication is considered to be the received vector, which is decoded to the codeword at the top of the column. We assume the enrollment and the probe vector in a genuine authentication to belong to the same column of the standard array which implies that they would decode to the same codeword at the top of the column in which case the hashes would match and access would be granted. Another important observation is that the enrollment and probe vector would have a different error pattern when compared to the codeword at the top of the column. This observation is important with regard to the noise included in the biometric during acquisition or processing and makes the system robust. The benefit of our approach is that we do not have to present any other side information to the decoder like a syndrome or a saved message key \cite{Sutcu_2008_Feature_SW,Sutcu_2008_Feature_ECC}.  

Our error control coding scheme can also be described in the context of fuzzy commitment. In fuzzy commitment, a random message is generated and encoded to give a n-tuple codeword which is combined (added) with the enrollment feature vector. The result of this combination is stored as a secure template. During authentication, the probe feature vector is combined with the stored template and decoded to get the estimate of the random message which is compared with the random message used during enrollment for authentication. In a standard-array context, the random message and its encoded codeword belong to the same column of the standard array. The enrolled biometric template is the coset-leader or the error pattern and when added to the codeword (of random message) and the probe, we get a template which when decoded gives us the estimate of the random message. For comparison purpose, we have also implemented our framework in the context of a fuzzy commitment scheme using RS codes and it gives the same result as with a secure sketch scheme.   
\vspace{-0.40cm}
\subsection{Enrollment and Authentication Procedure}
 The output of the DNN is the joint feature vector \textbf{e}. During enrollment, $G$ user-specific reliable components are selected from this vector \textbf{e} and the indices of these $G$ components form the user-specific key \textbf{g} . The feature vector \textbf{e} is binarized to generate a binary vector \textbf{a}. A new user-specific reliable bits vector $\textbf{r}_\textbf{a}$ is formed by selecting the bits from the binary vector \textbf{a} at the corresponding location or indices as specified by a key \textbf{g}. Reed-Solomon decoding is applied only on the reliable bits vector  $\textbf{r}_\textbf{a}$ to generate the multibiometric sketch $\textbf{s}_\textbf{a}$. The hash of this sketch $f_\mathsf{hash}$($\textbf{s}_\textbf{a}$)  is stored as a secure template in the database. The key \textbf{g} is securely stored in a location that is local to the matcher (i.e., it could be stored in the secure memory of a smartphone, or stored in the cloud if the matching is performed in the cloud). 

During authentication, the probe user presents the biometrics (face and iris) and the key \textbf{l} where \textbf{l} could be same as the enrollment key \textbf{g} in the case of genuine probe or it could be a different key for an impostor probe. Using the biometrics and the key \textbf{l}, the probe reliable bits vector $\textbf{r}_\textbf{b}$ is generated and decoded using a same RS code as in enrollment to generate an estimate of the enrolled multibiometric sketch $\hat{\textbf{s}}_\textbf{a}$. If the hash of the enrolled sketch $f_\mathsf{hash}$(\textbf{s}$_\textbf{a}$) matches the hash of the decoded estimate $f_\mathsf{hash}$($\hat{\textbf{s}}_\textbf{a}$), then the access is granted. In case hashes do not match, access is denied.

\section{EXPERIMENTS AND RESULTS}
\label{sec:exper}

In this section we will dicsuss the datasets and the experimental results. 

\begin{figure*}[t]
\centering     
\subfigure{\label{fig:a}\includegraphics[width=7.5cm]{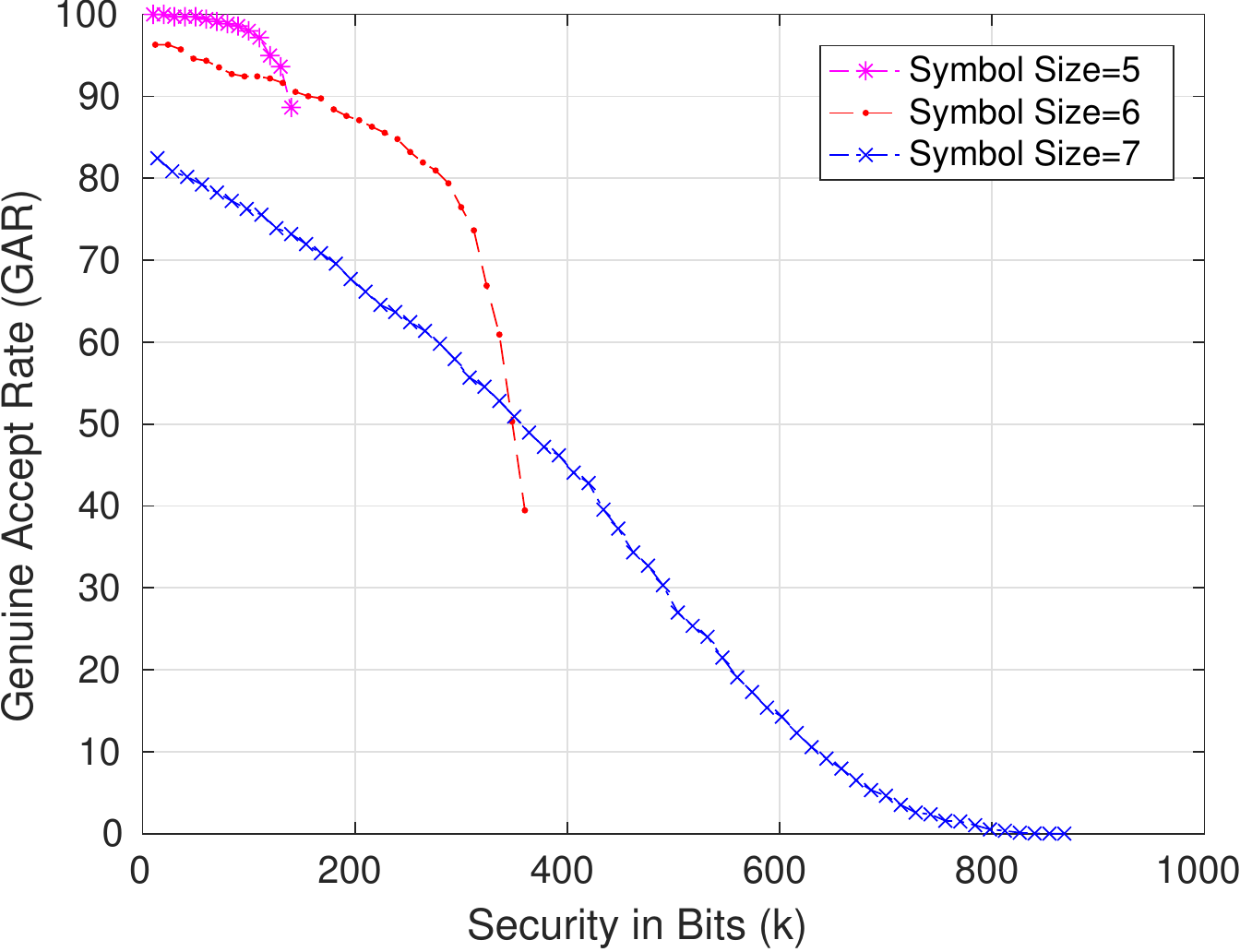}} \hspace{15mm} 
\subfigure{\label{fig:b}\includegraphics[width=7.5cm]{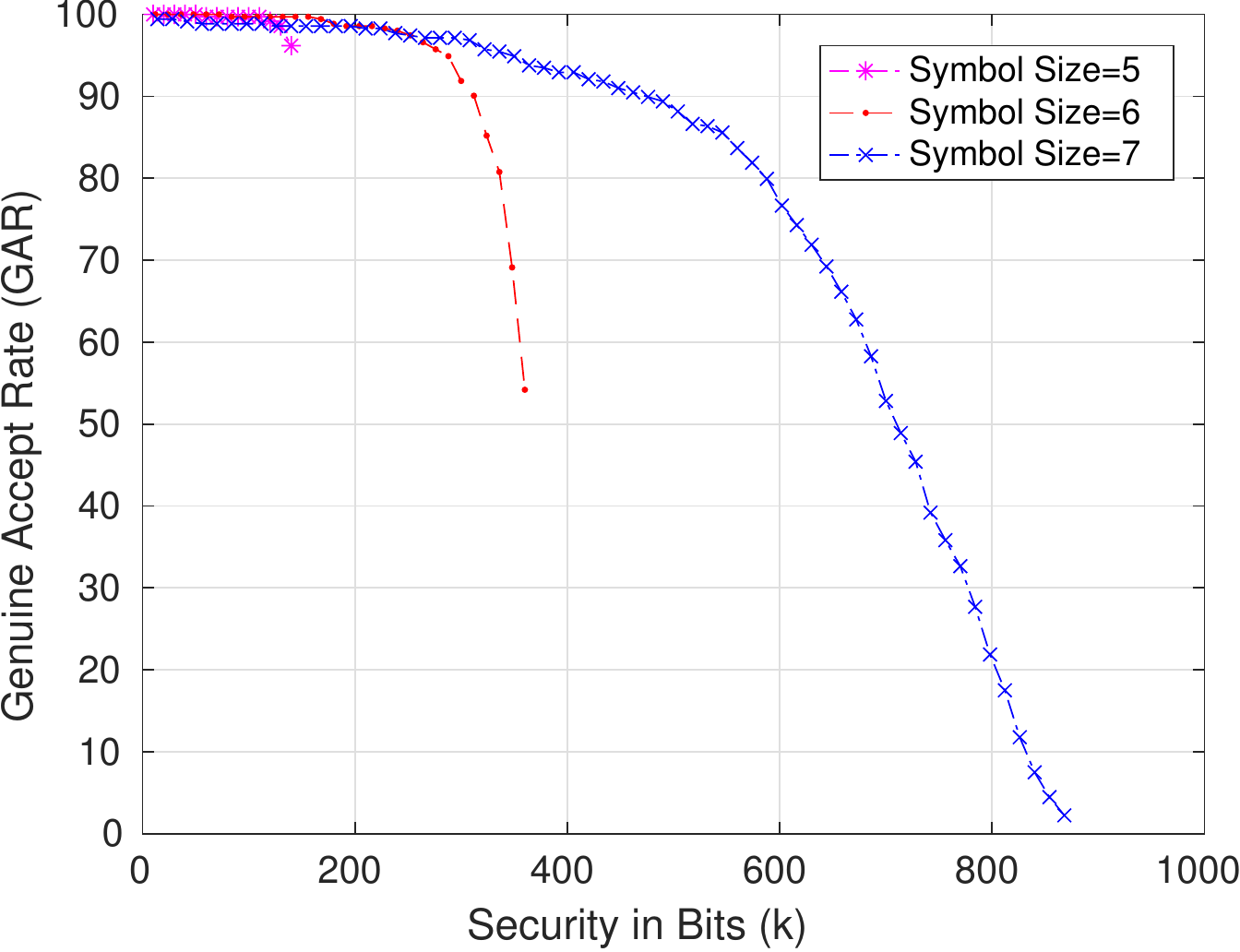}}
\caption{\comment{GAR vs Security curve for FCA (left) and BLA (right) using different symbol sizes.}}
\label{fig:roc}
\vspace{-0.30cm} 
\end{figure*}

\vspace{-0.40cm}
\subsection {\textbf{Datasets}}\label{subsec:DS}

The Face-CNN has been fine-tuned on Casia-Webface \cite{yi_learning_2014} which contains 494,414 facial images corresponding to 10,575 subjects. The Iris-CNN has been fine tuned on the combination of CASIA-Iris-Thousand\footnote[1]{http://biometrics.idealtest.org/} and ND-Iris-0405 \cite{bowyer_ndiris_2010} with about 84,000 images corresponding to 1355 subjects. The Face-CNN and Iris-CNN are also fine tuned on the 2013 face and iris subsets of the WVU-Multimodal 2012-2013 datasets\footnote[2]{http://biic.wvu.edu/} respectively. The WVU-Multimodal dataset for the year 2012 and 2013 together contains a total of 119,700 facial images and 257,800 iris images corresponding to 2263 subjects with 294 common subjects.
\vspace{-0.40cm}
\subsection {\textbf{Experimental Set Up and Results}}


\comment{The parameters used for fine-tuning the CNNs are similar to \cite{simonyan_very_deep_2014}, except that we have used a smaller batch size of 40}. For testing, we have used 50 subjects from WVU-Multimodal 2012 dataset and these 50 subjects have never been used in the training set. We randomly select 20 face and iris image pairs with no repetitions for each of these 50 testing subjects. These 1000 pairs $(50 \times 20 )$ are forward passed through the DNN and 4096-dimensional 1000 fused feature vectors are extracted. 

We then use the binarizing and reliable-bit extraction method from \cite{kevenaar_face_2005} to generate the cancelable multibiometric template. The number of reliable components selected from the fused feature vector depends on the number of bits per codeword in the RS code. The RS codes use symbols of length $m$. The codeword is of length $N= 2^{m-1}$ in symbols, which means the number of bits per codeword is $n=mN$. For example, if the $m=6$ then $N=63$ and $n=378$. In this case the number of reliable components chosen would be 378. This 378-dimensional vector is decoded to generate a multibiometric sketch whose length equals $km$ bits, where $k$ can be varied depending on the error correcting capability required for the code and $km$ also signifies the security of the system in bits. In our experiments we have taken $m=5,6,7,$ which means $n=155,378,889$ respectively.

An important metric to assess the performance of a biometric system is False Accept Rate (FAR) which is defined as the rate at which the system allows access to an unauthorized user. Considering the concept of standard array, it can be observed that the False Accept Rate (FAR) depends on the length of the multibiometric sketch $k$.  Cnsider a message length of $k$ bits, then the standard array has $2^k$ columns (one for each codeword). \comment{Empirically, it has been observed that} due to the binarization process, a impostor template will have bits that are equally likely to be 0's and 1's and therefore they are equally likely to be in any column of the standard array.  The probability (FAR) that they are in the same column as the enrollment template is thus $1/(2^k)$ or $2^{-k}$. We evaluate the trade-off between the recognition performance and the security of the proposed multibiometric protection scheme by using GAR-Security (G-S) curves. GAR is the Genuine Accept Rate and is defined as the rate at which the system allows access to an authorized or a genuine user. The G-S curve is obtained by varying the error correction capacity or the rate of the code (varying $k$ for the RS code with symbol size $m$). 

Fig. \ref{fig:roc} gives the G-S curves for FCA and BLA for different symbol size $m$. As we can observe from the plots, for both FCA and BLA, the best recognition performance is given for symbol size = 5 or 6 and we can get a GAR of at least $92.5\%$ for a security of 100 bits. We can also observe from Fig. \ref{fig:roc} that when we use a symbol size of 7 we can still get a good GAR of above 90\% for a security level of 400 bits.   Table \ref{table:FRR} summarizes the GAR of FCA and BLA for different symbol sizes at a minimum security level of 53 bits, which is equivalent to the guessing entropy of an 8-character password randomly chosen from a 94-character alphabet \cite{Burr_2004_Electronic}. From the table we can observe that BLA gives much better performance than FCA architecture for any symbol size.

An important observation here is that the number of reliable components chosen from the output of the DNN depends on the number of bits per codeword $n$, however, the security of the system is controlled by the error correcting capability or the rate of the code. For example, for a symbol size $m$ = 5 and $n$ = 155, to have a security ($k$) of 100 bits we need to use a RS code of rate $k/n=100/155=0.65$. 

\begin{table}[t]
\centering
\begin{tabular}{|c|p{0.4cm}|c|p{0.35cm}|c|c|}
 \hline
\multicolumn{1}{|c}{\multirow{2}{*}{Symbol Size}} &\multicolumn{1}{|c}{\multirow{2}{*}{$n$}} &\multicolumn{1}{|c}{\multirow{2}{*}{Security}} &\multicolumn{1}{|c}{\multirow{2}{*}{$k/n$}} &\multicolumn{2}{|c|}{GAR} \\ [0.5ex] 

 \cline{5-6}
($m$)& & ($k$) & & FCA & BLA \\ \hline \hline
\multirow{3}{*}{5} & \multirow{3}{*}{155} & 53 & 0.34 & 99.65\% & 99.85\% \\ \cline{3-6}
& & 80 & 0.52 & 98.86\% & 99.7\% \\ \cline{3-6} 
& & 100 & 0.65 & 98\% & 99.68\% \\ \hline \hline
\multirow{3}{*}{6} & \multirow{3}{*}{378} & 53 & 0.14 & 94.55\% & 99.99\% \\ \cline{3-6}
& & 80 & 0.21 & 93\% & 99.75\% \\ \cline{3-6} 
& & 100 & 0.26 & 92.5\% & 99.7\% \\ \hline \hline
\multirow{3}{*}{7} & \multirow{3}{*}{889} & 53 & 0.06 & 79.5\% & 98.9\% \\ \cline{3-6}
& & 80 & 0.09 & 77.5\% & 98.9\% \\ \cline{3-6} 
& & 100 & 0.11 & 76.2\% & 98.5\% \\ \hline


\end{tabular}
\caption{Genuine Accept Rates of FCA and BLA for a security level of 53, 80 and 100 bits using different symbol size. }
\label{table:FRR}
\vspace{-0.50cm}
\end{table}

\section{Security analysis}

We can measure the security of a biometric secure scheme in terms of privacy leakage for various scenarios of data exposure. The privacy leakage is quantified using mutual information as $I(\textbf{x};V)=H(\textbf{x})-H(\textbf{x}|V)$, where $V$ represents the loss of information and $\textbf{x}$ is the enrolled binary feature vector. The information can be lost when the key \textbf{g} and/or the sketch $\textbf{s}_\textbf{a}$ is compromised. H(\textbf{x}) quantifies the number of bits required to specify $\textbf{x}$ and $H(\textbf{x}|V)$ quantifies the remaining uncertainty about $\textbf{x}$ given knowledge of $V$, the mutual information $I(\textbf{x};V)$ is the reduction in uncertainty about $\textbf{x}$ given $V$ \cite{rane_secure_biom_2013}.

The dimension of the binary feature vector is 4096 which means the uncertainty $ H(\textbf{x})=4096$ because the bits of the binary feature vector are independent and identically distributed with maximum entropy \comment{due to the binarization process \cite{kevenaar_face_2005}}. If the adversary gains access to the key or the multibiometric sketch or both, the maximum reduction in uncertainty is given by $H(\textbf{x})-H(\textbf{x}|V) = 4096-n $ as the adversary has access to at most $n$ components (bits per codeword) using the key $\textbf{g}$ but not all the components of the feature vector $\textbf{x}$. Even if we consider $n=155$, it would still mean an uncertainty of $4096-155=3941$ bits which is computationally infeasible to crack with brute force attack. The security analysis shows that the generated multibiometric sketch is secure as the key is revocable and the error control coding and hashing provides the non-invertibility.   

\section{Conclusion}
The feature extraction capability of deep neural networks has been utilized to generate a shared joint representation by implementing two different fusion architectures, fully connected architecture and bilinear architecture. We have utilized the error correcting capabilities of a Reed-Solomon code to provide security in the form of a multibiometric sketch.  In our experiments, the multibiometric secure scheme has been tested using Reed Solomon codes of various symbol sizes.  Symbol size m=6 gives the best performance with a GAR as high  $99.7\%$ for a security of 100 bits, which is computationally infeasible to crack with brute force. We have shown that the classification performance can be improved for multiple modalities while maintaining good security and robustness. 

\vfill\pagebreak

\bibliographystyle{IEEE}
\bibliography{deep_multimodal}

\end{document}